# Characterizing the Influence of Features on Reading Difficulty Estimation for Non-native Readers

Yi-Ting Huang

Institute of Information Science

Academia Sinica

Email: ythuang@iis.sinica.edu.tw

Meng Chang Chen

Institute of Information Science

Academia Sinica

Email: mcc@iis.sinica.edu.tw

Yeali S. Sun

Department of Information Management

National Taiwan University

Email: sunny@ntu.edu.tw

**ABSTRACT**

In recent years, the number of people studying English as a second language (ESL) has surpassed the number of native speakers.  Recent work have demonstrated the success of providing personalized content based on reading difficulty, such as information retrieval and summarization. However, almost all prior studies of reading difficulty are designed for native speakers, rather than non-native readers. In this study, we investigate various features for ESL readers, by conducting a linear regression to estimate the reading level of English language sources. This estimation is based not only on the complexity of lexical and syntactic features, but also several novel concepts, including the age of word and grammar acquisition from several sources, word sense from WordNet, and the implicit relation between sentences. By employing Bayesian Information Criterion (BIC) to select the optimal model, we find that the combination of the number of words, the age of word acquisition and the height of the parsing tree generate better results than alternative competing models. Thus, our results show that proposed second language reading difficulty estimation outperforms other first language reading difficulty estimations.

**Introduction**

Reading difficulty (also called readability) is often used to estimate the reading level of a document, so that readers can choose appropriate material for their skill level. Heilman, Collins-Thompson, Callan, and Eskenazi (2007; denoted as the Heilman method hereafter in this paper) described reading difficulty as a function of mapping a document to a numerical value corresponding to a difficulty or grade level. A list of features extracted from the document usually acts as the inputs of this function, while one of the ordered difficulty grade levels is the output corresponding to a reader's reading skill.

The number of new documents and language material uploaded online is growing at a seemingly exponential pace. They are written in a various range of languages and for a wide range of readers. Despite this large volume of material however, several researchers, such as Collins-Thompson, Bennett, White, Chica, and Sontag (2011), note that current mainstream search engines poorly satisfy the needs of language learners. These researchers suggest the need for more personalized search results, based on reading proficiency of the user and the reading difficulty of relevant documents. Although such search methods may be largely unnecessary for the roughly 375 million native speakers of English (Curtis and Romney, 2006), there are now more than 200 million people (199 million－1.4 billion) taking English as a second language (ESL) around the world (English Language, n.d.). Wan, Li, and Xiao (2010) further noted that ESL reading is an important need for a large segment of the population, and subsequently proposed an easy-to-understand English summary algorithm based on reading level. This increasing degree of attention by researchers on this topic suggests two key areas for further research. First, regardless of the application type, such as information retrieval or summarization, providing readers with personalized search results based on reading difficulty is valuable. Secondly, the swelling numbers of ESL readers create an opportunity and need for more specialized application design.

Numerous researchers have studied the issue of reading difficulty, applying various lexical and grammatical features in statistic models. Early studies such as the Dale-Chall model (Dale and Chall, 1948), the Flesch-Kincaid measures (Kincaid, Fishburne, Rodgers, and Chissom, 1975), and the Lexile (Stenner, 1996) only adopted simple lexical features and designed a regression model to predict difficulty levels. Schwarm and Ostendorf (Schwarm and Ostendorf, 2005) proposed more complicated features that significantly increased the performance of readability prediction. Furthermore, some researchers also took language models into consideration (Collins-Thompson and Callan, 2004; Heilman et al., 2007; Heilman, Collins-Thompson, and Eskenazi, 2008), in order to obtain a probability distribution for each grade.

The majority of research on reading difficulty has focused on documents written for native speakers (also called first language), and comparatively little work (Heilman et al., 2007) has been done on the difficulties of documents written for non-native readers (also defined as second language). Non-native readers have a distinct way to acquire second language from native speakers. As Bates (2003) pointed out, there are wide differences in the learning timelines and processing times between native and non-native readers; native readers learn all grammar rules before formal education, whereas non-native readers learn grammatical structures and vocabulary simultaneously and incrementally. Almost all first-language reading difficulty estimations focus on vocabulary features, while second-language reading difficulty estimations especially emphasize grammatical difficulty (Heilman et al., 2007). Wan, Li and Xiao (2010) found that college students in China still have difficulty reading English documents written for native readers, even though they have learned English over a long period of time. These studies indicate that it is unsuitable to apply a first-language reading difficulty estimation directly; instead, second-language reading difficulty estimation must be developed.

Besides estimating the reading levels of documents, some researchers have applied reading difficulty estimations to analyze the quality of documents and textbooks (Kate, Luo, Patwardhan, Franz, Florian, Mooney, ... Welty, 2010; Agrawal, Gollapudi, Kannan, and Kenthapadi, 2011). In Agrawal et al. (2011), researchers implemented a readability analyzer for textbook writers to identify whether sections in textbooks are well written. Learning materials composed in textbooks are one of the effective approaches for guiding learners towards knowledge. A great number of textbooks are written by only a few domain experts. These experts write learning materials without an official grading standard, with their own estimated standard instead. Even though the difficulties of some learning materials increase gradually, various versions of learning materials could still be dissimilar. For example, in language learning textbooks, some words and grammar are taught at an easier level in one version of the textbook, while they are covered at a more difficult level in other versions. Thus, textbook writers may benefit significantly from using difficulty estimations when designing the context of their textbooks.

Starting from the idea that 1) non-native readers comprise a significant segment of the global population but their second-language reading skills are often poorer than native readers, and that 2) language textbook writers need an estimated standard to help them compose content, we propose a new reading difficulty model. This model is designed for non-native readers, and incorporates a large number of features. These features include prior variables selected from existing research, after testing their applicability for non-native learners, as well as several novel concepts first proposed in this research. These new features are briefly outlined below.

1. *Age of word acquisition*. The age of acquisition *(AOA)* of words represents when words are learned. Because the words learned by second language learners usually depend on the structure of the learning material, we employ grading indices of vocabulary as the age of word acquisition to estimate word difficulty.
2. *Grading index of grammar*. Heilman et al. (2007) found that a model with a complex syntactic grammatical feature set achieves more accurate results than models with simple grammatical syntax features. Hence, we collected 44 grammar patterns from textbooks as the age of grammatical acquisition. These represent grammar patterns that non-native readers have acquired at various grade levels.
3. *Word sense*. It is certain that a word has more than one sense or meaning. When given a word, its meaning may vary broadly depending on context. People may learn a particular meaning of a word at various ages. Thus, we designed semantic features to identify word senses in a document.
4. *Coreference relation*. Coreference is a grammatical relation that captures the implicit relations between sentences. For non-native readers, when they recognize a coreferent relation correctly, they may understand reading materials more clearly. Hence, we investigate coreferent relations when evaluating second-language reading difficulty.

Finally, in this paper we not only build a model to estimate reading difficulty, but also investigate the optimal combination of features for improving reading difficulty estimation. Our optimal model was selected by forward selection and judged by Bayesian Information Criterion (BIC). In our evaluation, we applied the model to a corpus of textbooks designed for ESL high school students in Taiwan. We examined the accuracy of the individual features and compared the proposed second-language reading difficulty estimation with alternative methods. Our results show that the age of word acquisition particularly plays an important role, and that the performance of the proposed reading difficulty estimation is better than alternative estimations.

The remainder of this study is organized as follows. Section 2 describes related work on reading difficulty. In Section 3, we define the research problem and present features of the task. Section 4 presents the reading difficulty model used in this study. Section 5 contains the experimental method and results. Finally, Section 6 provides discussion and Section 7 summaries our conclusions and offers ideas for future work.

**Related Work**

Early related work on testing reading difficulty only used a few simple features to measure lexical complexity, such as word frequency or the number of syllables per word. Because they took fewer features into account, most studies made assumptions on what variables affected readability, and then based their difficulty metrics on these assumptions. One example is the Dale-Chall model (Dale and Chall, 1948), which determined a list of 3,000 commonly known words and then used the percentage of rare words to measure lexical difficulty. Another example is the Lexile Framework (Stenner, 1996), which first used the mean log word frequency as a feature to measure lexical complexity. The researchers then entered the parameters into a logistic regression analysis to obtain a difficulty level, which helped determine if a reader could comprehend 75% of a given document. Using word frequency to measure lexical difficulty assumes that a more frequent word is easier for readers. Although this assumption seems fair, since a widely used word has a stronger chance to be seen and absorbed by readers, this method is susceptible to the diverse word frequency rates found in various corpora.

More recent approaches have started to take n-gram language models into consideration to assess lexical complexity, which can more accurately measure difficulty. Collins-Thompson and Callan (2004) used the smoothed unigram language model to measure the lexical difficulty of a given document. For each document, they generated language models by levels of readability, and then calculated likelihood ratios to assign the level of difficulty; in other words, the predicted value is the level with the highest likelihood ratio of the document. Similarly, Schwarm and Ostendorf (2005) also utilized statistical language models to classify documents based on reading difficulty level, and they found that trigram models are more accurate than bigrams and unigrams.

In addition to using fairly basic measures to calculate lexical complexity, prior studies also often only calculated the mean number of words per sentence to estimate grammatical readability. Using sentence length to measure grammatical difficulty assumes that a shorter sentence is syntactically simpler than a longer one, however long sentences are not always more difficult than shorter sentences. In response, more recent approaches have started to consider the structure of sentences when measuring grammatical complexity and making use of increasingly precise parser accuracy rates. This research usually considered more grammatical features such as parsing features per sentence in order to make a more accurate difficulty prediction. Schwarm and Ostendorf (2005) employed four grammatical features derived from syntactic parsers. These features included the mean parsing tree height, mean number of noun phrases, mean number of verb phrases, and mean number of subsidiary conjunctions to assess a document's readability. Similarly, Heilman et al. (2008) used grammatical features extracted from an automatic context-free grammar parsing trees of sentences, and then computed the relative frequencies of partial syntactic derivations. In their model, the more frequent sub trees are viewed as less difficult for readers.

As mentioned above, almost all past literature was designed for native readers, and this literature consulted word frequency from general corpora that were composed of articles written for native readers. But for non-native readers, the word difficulty depends on the structure of the material they study, not its popularity in the real world. In this study, we design a reading difficulty estimation for a given document for non-native readers. We investigate the effectiveness of several meaningful lexical and grammatical features from early work, and then further consider organized grading indices of vocabulary from different sources, as well as grammar patterns collected from textbooks—those which represent words and grammar patterns that non-native readers have acquired at various grade levels, such as the age of a word and grammatical acquisition. Furthermore, we also propose features that take into consideration word sense and coreference resolution. In our evaluation, ESL readers serve as an empirical example to represent non-native readers, and documents were estimated for them. From this setup, experiments were conducted to evaluate the importance of each feature for reading difficulty estimation, and then these features were further investigated to determine their most optimal combination.

**Reading Difficulty Estimation**

Our approach selects several representative features to generate a function to predict reading difficulty. The inputs of this function are a list of features of a given document, and the output is a difficulty score for the document. The scores can also correspond to one of the ordered difficulty levels.

Let *D* represents a document, while *S* represents the sentences in *D*. Suppose that *D* has *n* sentences, $s_1, s_2, ..., s_n$, so that $D = \{ s_1, s_2, ..., s_n \}$. Let *W* be the set of words in *D*. Suppose *D* has *m* distinct words, $w_1, w_2, ..., w_m$, so that a document $D = \{ w_1, w_2, ..., w_m \}$. We further suppose that the sentence *S* has *k* words, $w_1, w_2, ..., w_k$, so that $S = \{ w_1, w_2, ..., w_k \}$, $m > k$. For a given training data set, the features are extracted and sent to a linear regression process to obtain a linear model that includes the weight of each feature. The linear model is then applied to a document to estimate the difficulty level. In the following sections we explain and define the features used in the proposed estimation.

**Baseline Features**

*Word Number:* A basic assumption is that a longer document is more difficult than a shorter one. Almost all prior work assumed that the number of words in a document accurately estimates reading difficulty (Flesch, 1948; Dale and Chall, 1948; Gunning, 1952; McLaughlin, 1969; Coleman and Liau, 1975; Kincaid et al., 1975). Pitler and Nenkova (2008) pointed out that this feature is significantly correlated with readability. For non-native readers, a longer document takes more time to comsume. Therefore, the number of words in a document is used in this study as one of the features to estimate reading difficulty. Word count difficulty is defined as follows:

$$word\_number = \log |D|$$

*Sentence length:* Past studies have also taken sentence length into account, assuming that a shorter sentence is easier than a longer one (Flesch, 1948; Dale and Chall, 1948; Gunning, 1952; McLaughlin, 1969; Coleman and Liau, 1975; Kincaid et al., 1975). Thus for each document, we consider the average number of words per sentence as sentence length. The sentence difficulty is defined as follows:

$$sentence\_length = \frac{word\_number}{n}$$

*Syllables:* A syllable is a unit of organization for a sequence of speech sounds. For example, the word *water* is composed of two syllables: *wa* and *ter*. Some related work has also taken syllables into consideration (Flesch, 1948; Gunning, 1952; McLaughlin, 1969; Kincaid et al., 1975). One notable example is the SMOG formula (McLaughlin, 1969), which estimates the reading difficulty of a document by only using the average number of polysyllables per senence.

Even though syllables have proven to be a useful measure of reading difficult for first-language users, similarities between sounds of a native speaker's mother tongue and their adopted second language can impact second-language learning. For instance, a word in an Asian language usually has one syllable, while a word in western languages usually has more than one syllables. When learning a second language, a second-language learner could use similar-sounding syllables from their first language to learn vocabulary (called L1－phonology effect hypothesis; Yamada, 2004). We assume the number of syllables in a word may affect the difficulties of documents. Thus, we find the average number of syllables of every word in a document to measure reading difficulty. The syllable difficulty of a document is defined as follows:

$$syllables = \frac{\sum_{i=0}^{m} word\_syllables_i}{m}$$

where word_syllables$_i$ is the number of syllables within a word *i*.

**Age of Acquisition Features**

We adopt organized word lists for determining the age of word acqusition. Chall and Dale (1948) counted the number of words not in a list of the most basic 3,000 words to estimate vocabulary difficulty. Kidwell, Lebanon, and Collins-Thompson (2009) revealed that word frequency described as an age distribution could infer reading difficulty. These studies note reading difficulties related with not only documents but also readers' knowledge.

It is crucial to understand when a word is acquired by target readers. Kireyev and Landauer (2011) have tried using latent semantic analysis to capture word difficulty. Even though there is no existing non-native dictionary presenting the age of word acquisition, non-native readers learn vocabulary in a limited range, which is usually decided by experts or teachers. Similar to Wan et al. (2010), we build two dictionaries from educational grading indices made by human experts. This helps better identify the vocabulary difficulty level of non-native readers. Two resources, the General English Proficiency Test Reference Vocabulary and the Vocabulary Quotient, are used to estimate the age of word acquisition in our study.

*GEPT Word Lists:* The General English Proficiency Test (GEPT; Wu and Liao, 2010) is designed to evaluate student proficiency in English as a second lanaguage. It provides a reference vocabulary list with about 8,000 words divided into three word levels: elementary (gept1), intermediate (gept2) and high-intermediate (gept3). Some words not found in the GEPT word list are attributed to the out of GEPT word list (gept0). For each word from a document, we identify its vocabulary difficulty by searching for the word's level from the GEPT word lists, counting the number of distinct words in each level, and finally normalizing by the total number of distinct words in each level.

*Age of Word Acquisition:* In addition to the GEPT, we also collected a word list from an orgaization, Vocabulary Quotient (VQ; Ho and Huong, 2011). This organization collected more than 10,000 words and labeled them in reference to other educational institutions, such as the Elementary School Reference Vocabulary and the Junior High School English Reference Vocabulary texts made by the Minstry of Education of Taiwan, and the High School English Reference vocabulary text made by the College Entrance Examination Center of Taiwan. The word list is divided into fourteen levels (vq3 －vq16), which represent the words learned by non-native readers from elementary school to college. Just as with the GEPT list, some words are still absent from the Vocabulary Quotient word list; these words are attributed to out of vocabulary list (vq0). For each word from a document, we identify its difficulty by first referencing its difficulty level from within those word lists, and after counting the number of distinct words in each level, normalizing by the total number of distinct words in each level.

**Frequency Features**

Besides the age of word acquisition, word frequency is another approach to estimating word difficulty. Word frequency is based on the assumption that more frequent words are easier to identify. In our study, for every word in a document, we find its word frequency from the BNC corpus and also use a Google search result count as an alternative frequency.

*Word Frequency in BNC Corpus:* The British National Corpus (BNC; Lou and Guy, 1998) is a 100 million word collection of written and spoken language from a wide range of sources, designed to represent a wide cross-section of British English from the later 20th century. For each word in a document, we calculate the distinct word frequency (*wf*) that refers to the times it appears in the BNC corpus. Word frequency is defined as follows:

$$wf_i = \frac{n_i}{|d_j|}$$

where $n_i$ is the number of occurrences of the considered distinct word $w_i$ in document $d_j$, and the denominator is the sum of the number of occurrences of all distinct words in document $d_j$, that is, the size of the document $|d_j|$. For each word in a given doucment, we also calculate the average number of log word frequency. The document's difficulty value based on word frequency in the BNC corpus is defined as follows:

$$bnc\_freqency = \log \frac{\sum_{i=0}^{m} wf_i}{m}$$

*Google Search Result Count:* For a given query, Google will return a list of documents containing the queried words and a search result count. We use the search result count as a measure of word frequency, like the word frequency from a corpus. For each word in a given doucment, we also calculate the average number of log word frequency. The document's difficulty value based on word frequency from Google is defined as follows:

$$google\_search\_count = \log \frac{\sum_{i=0}^{m} google_i}{m}$$

where *google<sub>i</sub>* is the search result count of a word *i* from Google.

**Parsing Features**

Syntactic constructions affect the understanding of a sentence. This assumes that the more complicated a sentence, the greater its difficulty. Schwarm and Ostendorf (2005) proposed four syntactic features for their measure of reading difficulty: the average parsing tree height, the average number of noun phrases per sentence, the average number of verb phrases per sentence, and the average number of subordinate clauses per sentence (SBAR). Because sentences with multiple noun phrases require the reader to remember more entities, Barzilay and Lapata (2008) found that documents written for adults tended to contain more noun phrases than those written for children. In addition, while including more verb phrases in each sentence increases sentence complexity, adults might prefer to have related clauses explicitly grouped together. Pitler and Nenkova (2008) have also found a strong correlation between readability and the number of verb phrases. These works show that the more complicated the parsing features in a document, the more likely it was written for adults. Hence, we also examine the influence of parsing features for non-native readers.

Prepositions are a class of words that indicate relationships between nouns, pronouns and other words in a sentence. Prepositions can be divided into two kinds: simple prepositions and compound prepositions. Simple prepositions are single word prepositions, while compound prepositions are more than one word. We assume that more prepositional phrases in a sentence also increases its complexity, and non-native readers might be confused by complex prepositional phrases. Thus, in addition to the parsing features proposed by Schwarm and Ostendorf (2005), we also present the average number of prepositional phrases as a new feature to capture grammatical complexity.

Thus, from the outline above, for a document we consider the following syntactic features from parsing results generated by a stanford parser (Klein and Manning, 2003): the average parsing tree height, the average number of noun phrases, the average number of verb phrases, the average number of SBAR and the average number of prepositional phrases.

*Average Parsing Tree Height:* Suppose the height of a parsing tree of a sentence is *h*. The average parsing tree height difficulty of a document is defined as follows:

$$tree\_height = \frac{\sum_{i=0}^{n} h_i}{n}$$

*Average Number of Noun Phrases:* Suppose a sentence has $np_i$ noun phrases. The average noun phrase difficulty of a document is defined as follows:

$$np = \frac{\sum_{i=0}^{n} np_i}{n}$$

*Average Number of Verb Phrases:* Suppose a sentence has $vp_i$ verb phrases. The average verb phrase difficulty of a document is defined as follows:

$$vp = \frac{\sum_{i=0}^{n} vp_i}{n}$$

*Average Number of SBAR:* Subsidiary conjunctions (SBAR), for example, *because*, *unless*, *even though*, and *until*, are placed at the beginning of a subordinate clause that links the subordinate clause and the dominant clause. SBAR is an indicator to measure sentence complexity. The SBAR difficulty of a document is defined as follows:

$$sbar = \frac{\sum_{i=0}^{n} sbar_i}{n}$$

*Average Number of Prepositional Phrases:* Suppose a sentence has $pp_i$ prepositional phrases. The average number of the prepositional difficulty of a document is defined as follows:

$$pp = \frac{\sum_{i=0}^{n} pp_i}{n}$$

**Grammar Features**

In Heilman et al. (2007), they found that grammatical features played an important role in reading difficulty estimation for second-language learners. A model with complex syntactic grammatical feature sets achieved more accurate results than simpler models. In their work, they examined the ratio of grammatical occurrence per 100 words: both the passive voice and past participle had obvious differences between the lowest and highest levels in the second-language corpus. Thus, we measure grammatical difficulty as a linguistic processing factor in estimating reading difficulty for non-native readers.

*Grading Index of Grammar (grammar1 −grammar6):* To decide the grammatical difficulty level of a document, we first collected sentences from the six versions of second-language textbooks and parsed the sentences to find their grammar patterns, for a total of 44 grammar patterns. Manually identifying these grammar patterns allows the parsing tool to then automatically find these same patterns within a given document. Next, using this parsing tree structure searching tool (Levy and Andrew, 2006), the grammatical structures were assigned to the textbook grade in which they frequently appear.

**Semantic Features**

For any given word, its meaning may vary broadly depending on the context. For example, the word *bank* has two distinct meanings (also called two senses), *financial institution* and *sloping mound,* not to mention its other collquial uses. For both the age of word acquisition and frequency features, we assume that a word only has one sense, because this still results in accurate performance with many language technologies, such as information retrieval or text classification. However, it cannot be claimed that a non-native reader having learned a word knows every sense of the word. Therefore, we designed semantic features to identify word senses in a document.

*Average Number of WordNet Synsets:* We adopted WordNet (Miller, Beckwith, Fellbaum, Gross, Miller, 1990) as a resource for understanding the senses in a word. WordNet is a large lexical database of English. The database contains 155,287 words, with each word annotated with a set of senses. The average noun has 1.23 senses and the average verb has 2.16 senses. A set of near-synonyms is defined as a synset, which represents a concept of a word.

For each word in a document, we total the number of a word's synset using WordNet. To determine the representation of this feature, we develop seven categories (wordnet1－wordnet7) to represent the number of synsets of each word in a document. Here, suppose a word has $ws_i$ synsets. The number is normalized as two square roots and then rounded down to an integer as a feature index. For example, if the number of synsets of a word is 17, it is attibuted to wordnet4. If the number of synsets of a word is greater than 49, it is assigned to wordnet7. Finally, we count the number of distinct words in each WordNet category and normalize by the total number of distinct words.

**Coreference Features**

Coreference is a gammatical relation that presents two referring expressions that refer to the same entity. This entity is called an antecdent, and the referring expression is called an anaphora. We assume that coreference represents the implicit relations between sentences. When non-native readers recognize the coreferent relation well, they might be able to understand the reading material more clearly. For a document, we count the number of pronouns per document, the number of proper nouns per document, the number of antecedents per document, the average number of anaphora per coreference chain and the average distance between anaphora and antecedents per chains.

*Average Number of Pronouns:* We assume that the greater the number of pronouns in a document, the more entities the reader needs to remember, and this increases reading difficulty. Thus, we total the average number of pronoun in a doucment.

*Average Number of Proper Nouns:* If a sentence contains more than one proper noun, a reader must remember more objects in a document. Barzilay and Lapata (2008) found that documents written for adults tended to contain more entities than those written for children. Hence, we count the average number of proper nouns in a document.

*The Number of Antecedents per Document:* Antecedents represent real entities mentioned in the document. Similar to the average number of proper nouns, we assume that if a document contains less entities, the document is easier to read. We total the number of antecedents as the number of enities to capture this idea.

*The Average Number of Anaphora per Corference Chain (corefer_chain):* We assume that with more anaphora per coreference chain, non-native readers need more knowledge to resolve them; consequently, we count the average number of anaphora per chain.

*The Average Distance between Anaphora and Antecedents per Chain (corefer_distance):* This captures the distance between antecedents and anaphora. We assume if an antecedent and anaphora are in the same sentence, the sentence will be easy to understand. In constrast, if they are several sentences apart, it is probable that the document is more complex to read.

**Regression Model**

Linear regression is an approach to modeling the relationship between a scalar variable *Y* and variables denoted *X*. A prediction of a given document is the inner product of a vector of feature values for the document and a vector of regression coefficients estimated from the training data.

$$Y = \alpha + \sum_{i=1}^{n} \beta_i X_i + \varepsilon, \ i = 1, 2, ... n$$

where *Y* is the difficulty value of a document, $\alpha$ is the intercept parameter, $X = \{x_1, x_2, ..., x_n\}$ represents the feature values, $\beta = \{\beta_1, \beta_2, ..., \beta_n\}$ refers to the regression coefficient for each feature value *i*, and lastly $\varepsilon$ is an unobserved random variable that represents noise in the linear relationship between the dependent variable and regressors.

The primary reason for adopting linear regression for a reading difficulty model is that the output scores are continuous and related with each other, whereas the outputs of other methods, such as classification, are discrete and unrelated between levels. Kate et al. (2010) evaluated the performance of reading difficulty among several machine learning methods and reported that all of the regression family outperformed the baseline. In our study, the readability of a text is represented by a score or a class, which is typically indicated in terms of school grades. Overall, the content difficulty of textbooks increases incrementally. Thus, we opt for linear regression as our model, as we assert that our estimated results are correlated.

**Experiments**

In this section, the proposed features and model are evaluated. To determine how each feature contributes to an accurate readability judgment, we first conducted an experiment to test the performance of each feature using linear regression algorithms. Next, an optimal features set is determined by model selection, in order to investigate how to best combine the features that improve reading difficulty estimation. Finally, the proposed estimation was also modeled as a multiclass classification and compared to other related work. These experiments are described in the following subsections.

**Data for Evaluation**

Our experiment used data from senior high school English textbooks designed for Chinese students in Taiwan to learn English as a second language. We gathered *342* documents from five different publishers (including The National Institute for Compilation and Translation, Far East Book Company, Lungteng Cultural Company, San Min Book Company, and Nan-I Publishing Company). Poets and scripts were excluded from the data set because their formats are different from normal articles. Each document was graded using six levels, ranging from one to six. These levels indicate the semester grade levels of senior high school students, and were used as the gold standard in this work.

During data preprocessing, a majority of words were adopted directly and stop words were also used. A document consists of different forms of words, such as *promise*, *promises*, *promised*, *promised*, and *promising*. Even though they are derivative of the same root, they are acquired and used in different ages. For example, in our scenario, the word *promise* is taught in the second semester, while the word *promising* is in the fourth semester. This suggests that the different forms of words are represented as different meanings. Thus, initially, words were used directly in our study; otherwise, some words were lemmatized when necessary. For the same reasons, stop words are retained in the proposed estimation except for frequency features.

Five-fold cross-validation was employed. The data was first split into five sets. One set was used as held-out data to predict the reading difficulties of documents, while the rest of the data was used as training data to build a regression model. Each fold was further repeated five times by changing the pairing of training and testing.

**Metrics for Evaluation**

To evaluate the effectiveness of the estimated reading difficulty, the root mean squared error (RMSE) and Pearson's correlation coefficient (*r*) were used in the experiments. RMSE measures the averaged erroneous value between ground truths and estimated responses. It is an averaged distance for measuring how far estimated responses approach ground truths; the less the RMSE, the better the estimation. The Pearson's correlation coefficient ($r$) measures the trends between the ground truth and the generated results. It represents the strength of the linear relationship between two random variables. A high correlation shows that simple documents are estimated as having a low difficulty value, while difficult documents are predicted as having a high difficulty value.

**Evaluation of the Features**

First, an evaluation was designed to understand the impact of features. Table 1 summarizes RMSE and the correlations among different feature categories. The rows of a block represent the different feature categories in Section 3. The baseline features have the best results, leading to an RMSE of *0.98* and correlation of *0.82*. This was composed of word number, sentence length, and syllables, and indicates that they represent how a majority of experts design learning materials.

The second and third feature categories were GEPT and VQ, both of which are age of word acquisition features. They represent when non-native readers learn words at different ages. It is highly likely that age of word acquisition is also an important factor in analyzing reading difficulty for non-native readers. Even though GEPT and VQ features are derived from different resources and classified by fine-coarse grade respectively, both of their performances were high correlated with ground truth and the RMSE values were less or equal to *1.25*. Surprisingly, the GEPT word list, only divided into three levels, performs better than VQ, which categorizes words acquired by non-native readers from elementary schools to universities.

The fourth feature category was coreference features. The coreference features captured the inter-relationship between noun phrases. While baseline, GEPT, and VQ features only took explicit words into consideration, coreference features considered not only noun phrase types but also implicit interaction between noun phrases. Feng, Jansche, Huenerfauth, and Elhadad (2010) first proposed coreference and discourse features. In their investigation, noun phrase features and coreference inference features slightly improved. The result was fairly consistent in the second-language materials.

The fifth and sixth features were parsing and grammar features. Both features analyze sentence structures in a document. While parsing features were automatically extracted from a parser, grammar features were identified from a tree structure search tool based on manual grammatical patterns. The results of our grammar features were consistent with Heilman et al. (2007). In their study, vocabulary-based features produced more accurate results than grammar-based features alone, but complex grammar features performed better than simple ones. Even though we collected more than forty grammatical patterns from six grades in textbooks (more than the Heilman method), the results indicated that parsing features were slightly better than grammar features. It is possible that parsing features could be more robust than grammar features.

The next features are semantic features. Unfortunately, calculating the number of senses for each word seemed to have little impact on reading difficulty estimation. No matter how many senses a word has, the most important factor is whether the readers understands the specific sense of words in a document or not. The better solution might first determine each sense of word in a document, and then assess when reader had learned those meanings. This may represent a new research problem for future studies.

The last remaining features were frequency features derived from the BNC corpus and Google search engine. Surprisingly, frequency features were not good indicators for estimating reading difficulty. This went against Tanaka-Ishii, Tezuka, Terada (2010), who used the log frequency obtained from corpora as features to predict document reading difficulty. One explanation for this is that the format of features and the method in (Tanaka-Ishii et al., 2010) may be very different from this study. Another explanation is the possibility that lower and higher word frequencies are counteracted by the summarization of word frequencies.

Table 1: Results of RMSE and correlation among different feature categories.

| Categories | Features | RMSE | $r$ |
| --- | --- | --- | --- |
| Baseline | baseline-only | 0.98 | 0.82 |
| AOA | gept-only | 1.24 | 0.68 |
| AOA | vq-only | 1.25 | 0.67 |
| Coreference | coreference-only | 1.48 | 0.48 |
| Parsing | parsing-only | 1.50 | 0.46 |
| Grammar | grammar-only | 1.61 | 0.31 |
| Semantic | wordnet-only | 1.60 | 0.33 |
| Frequency | bnc_freqency | 2.22 | 0.12 |
| Frequency | google_search_count | 4.50 | -0.04 |

To evaluate the predicted performance in more detail, the results of each individual feature are shown in Table 2, in decreasing order of importance. The best performing feature was word number. Non-native readers have greater difficulty with longer documents, and this phenomenon might increase the non-native readers' psychological burdens. When experts compose the content of textbooks, they intuitively design shorter documents in lower grade textbooks and longer documents in higher-grade textbooks.

Second, the fourth and fifth best features were GEPT features. As with the feature categories in the previous subsection, the GEPT features were the second-best feature category to estimate reading difficulty. The gept1 word list refers to the organized word list that should be acquired from elementary school to senior high school. As seen in Table 2, the coefficient value of the gept1 was negative. It is very probable that a document having large portions of words from elementary schools and junior high schools might lead to lower grades. The gept2 and the gept3 were intermediate and high-intermediate levels in The General English Proficiency Test (GEPT). These word lists refer to the organized words that should be acquired during junior high school. The coefficient values of the gept2 and the gept3 were all positive. This suggests that documents with large portions of words from senior high school benefit from the grading of word difficulty. The performance of the gept3 was slightly better than that of the gept2. The words in the gept3 lists were also more challenging than those in gept2. Thus, it is possible that the more challenging word lists contributed to the reading difficulty estimation.

The top third feature was vq12. This was an organized word list that non-native readers acquired in the third grade in senior high school. As shown in Table 2, the coefficient values of vq12 are positive. As in the previous paragraph, this result also supports our finding that more challenging word lists led to more accurate reading difficulty estimations. The age of word acquisition derived from VQ divided words into finer levels. Compared with the word categories designed by GEPT, vq12 performed better than that of gept2 and gept3; six of the top twelve features were from vq categories, including vq12, vq5, vq10, vq11, vq4 and vq13. Except for word number, syllable and noun phrases, almost all of the top twelve features were related to age of word acquisition. This finding suggests that reading difficulty estimations should consider information not only from the document but also from specific readers.

The noun phrases feature (np) is in the top ten results of individual features, and it is also the first parsing feature in the decreasing performance list of Table 2. This finding is consistent with previous work (Pitler and Nenkova, 2009; Barzilay and Lapata, 2008). For non-native readers, the more noun phrases within in a document, the more entities that need to be remembered, and this impacts reading difficulty estimation. Moreover, the parsing features ranked from within the top 22 to top 10 of all features. Thus, in addition to word acquisition, these results suggest that parsing features could also be good indicators to estimate reading difficulty.

The remaining features were almost entirely coreference features, semantic features, grammar features and frequency features. The decreasing order of the individual features is similar to that of the feature categories in Table 1. Among the coreference features, the coreference distance feature performs the best. Although prior studies have used coreference features (Feng et al., 2010), our study is the first to note the significance of individual coreference features. When antecedents are far from the location of anaphors, it is not easy for non-native readers to quickly understand the interaction between noun phrases among sentences. Moreover, the first semantic feature in Table 2, wordnet3, performed better than the first grammar feature, grammar4. This result was different from the above-mentioned performance of feature categories in Table 1. Furthermore, the coefficient values increased from

wordnet1 to wordnet7, which supports our assumption that a document containing mostly words consisting of only a few senses is likely easier to read.

Table 2: Result of RMSE and correlation among individual features.

| Rank | Categories | Name | Regression | RMSE | r |
|---|---|---|---|---|---|
| 1 | baseline | word_number | y=0 * word_number + 0.47 | 1.25 | 0.67 |
| 2 | AOA | gept1 | y=-16.03 *gept1 +15.38 | 1.28 | 0.65 |
| 3 | AOA | vq12 | y=61.88 * vq12 + 1.43 | 1.34 | 0.61 |
| 4 | AOA | gept3 | y=53.59 * gept3 +1.31 | 1.35 | 0.60 |
| 5 | AOA | gept2 | y=27.9 * gept2－0.027 | 1.38 | 0.58 |
| 6 | AOA | vq5 | y=-19.81 * vq5 + 8.66 | 1.40 | 0.56 |
| 7 | baseline | syllables | y=6.15 * syllables－6.47 | 1.43 | 0.54 |
| 8 | AOA | vq10 | y=35.18 * vq10 + 0.96 | 1.46 | 0.51 |
| 9 | AOA | vq11 | y=54.75 * vq11 + 2.01 | 1.46 | 0.50 |
| 10 | parsing | Np | y=0.55 * np + 0.41 | 1.50 | 0.46 |
| 11 | AOA | vq4 | y=-28.61 * vq4 + 6.74 | 1.51 | 0.45 |
| 12 | AOA | vq13 | y=98.84 * vq13 + 2.63 | 1.51 | 0.45 |
| 13 | parsing | tree_height | y=0.57 * tree_height－2.04 | 1.52 | 0.44 |
| 14 | coreference | corefer_distance | y=0.01 * corefer_distance + 2 | 1.53 | 0.42 |
| 15 | parsing | Pp | y=1.18 * pp + 1.54 | 1.54 | 0.41 |
| 16 | parsing | Vp | y=0.76 * vp + 0.86 | 1.57 | 0.37 |
| 17 | AOA | vq6 | y=-22.93 * vq6 + 6.68 | 1.58 | 0.36 |
| 18 | AOA | gept0 | y=17.9582 * gept0 +1.8141 | 1.58 | 0.36 |
| 19 | AOA | vq3 | y=-21.61 * vq3 + 5.86 | 1.58 | 0.35 |
| 20 | baseline | sentence_length | y=0.03 * sentence_length + 2.27 | 1.60 | 0.33 |
| 21 | semantic | wordnet3 | y=-15.47 * wordnet3 + 6.2 | 1.61 | 0.31 |
| 22 | parsing | sbar | y=1.99 * sbar + 2.2 | 1.61 | 0.30 |
| 23 | AOA | vq9 | y=29.08 * vq9 + 1.53 | 1.61 | 0.30 |
| 24 | wordnet | wordnet1 | y=10.44 * wordnet1－0.79 | 1.61 | 0.30 |
| 25 | AOA | vq7 | y=34.19 * vq7 + 2.26 | 1.62 | 0.28 |
| 26 | grammar | grammar4 | y=3.59 * grammar4 + 2.39 | 1.64 | 0.23 |
| 27 | AOA | vq15 | y=240.02 * vq15 + 3.19 | 1.65 | 0.24 |
| 28 | coreference | antecedent | y=3.59 * antecedent + 1.17 | 1.66 | 0.2 |
| 29 | AOA | vq16 | y=206.58 * vq16 + 3.28 | 1.66 | 0.18 |
| 30 | semantic | wordnet4 | y=-25.19 * wordnet4 + 4.35 | 1.66 | 0.17 |

| 31 | semantic | wordnet5 | y=-31.1 * wordnet5 + 3.99 | 1.67 | 0.17 |
| 32 | grammar | grammar6 | y=1.99 * grammar6 + 2.99 | 1.68 | 0.13 |
| 33 | grammar | grammar3 | y=2.28 * grammar3 + 3.04 | 1.68 | 0.12 |
| 34 | grammar | grammar5 | y=2.15 * grammar5 + 2.95 | 1.68 | 0.13 |
| 35 | AOA | vq8 | y=21.63 * vq8 + 2.66 | 1.68 | 0.12 |
| 36 | semantic | wordnet6 | y=-25.25 * wordnet6 + 3.83 | 1.69 | 0.08 |
| 37 | semantic | wordnet7 | y=-27.19 * wordnet7 + 3.58 | 1.69 | 0.03 |
| 38 | grammar | grammar1 | y=0.39 * grammar1 + 3.22 | 1.69 | 0.00 |
| 39 | semantic | wordnet2 | y=2.34 * wordnet2 + 2.65 | 1.70 | -0.04 |
| 40 | grammar | grammar2 | y=0.25 * G2 + 3.36 | 1.70 | -0.06 |
| 41 | coreference | proper_noun | y=-1.15 * name_entity + 3.55 | 1.70 | -0.06 |
| 42 | coreference | pronoun | y=-2.32 * pronnoun + 3.66 | 1.70 | -0.08 |
| 43 | AOA | vq14 | y=-0.89 * vq14 + 3.47 | 1.70 | -0.16 |
| 44 | coreference | corefer_chain | y=-0.05 * corefer_chain + 3.65 | 1.70 | -0.14 |
| 45 | AOA | vq_null | y=-0.92 * vq0 +12.32 | 2.22 | 0.12 |
| 46 | freqency | bnc_freqency | y=-0.92 * bnc_frequency + 12.32 | 2.22 | 0.12 |
| 47 | freqency | google_result_count | y=-0.17 * google_result_count + 6.96 | 4.50 | -0.04 |

**Optimal Model Selection**

To investigate how combining features improves reading difficulty estimation, the forward selection were used to select the best subset of features for linear regression, and Bayesian information criterion (BIC; Schwarz, 1978) were applied to decide the best regression model. If a regression model employs every available feature (47 in total), it becomes sensitive to training data. In contrast, if a model was not designed well, its performance with the testing data should be poor. This section examines how this study identified an appropriate model with features that play important roles in determining reading difficulty.

The forward selection was employed to evaluate the optimal model; it starts with the intercept and adds at each step the features that most improve. The detailed rules for this process are as follows:

**Step 1.** The first feature with the highest Pearson Correlation Coefficient value was selected to the best model.

**Step 2.** The next selected feature had the highest semi-partial correlation, and is added into the model.

**Step 3.** After adding the new feature in step 2, the squared multiple correlation coefficient of new the model was calculated ($R^2$).

**Step 4.** To test whether the new feature contributes significantly to the model, the difference between the new $R^2$ value and old $R^2$ value was evaluated using Analysis of Variance (ANOVA).

**Step 5.** If the incremental difference in Step 4 significantly improved, the new feature stayed in the model; otherwise it was removed.

This process from step 2 to step 5 is repeated until the addition of further features produces no significant improvement.

Bayesian information criterion were used to select the best model based on estimating the Kullback-Leibler divergence between a true model and a proposed model, incorporating sample size. This process also introduces a penalty term for the number of parameters in a model. BIC is denoted as:

$$BIC = n \times \ln(\frac{RSS}{n}) + \ln(n) \times k$$

where *RSS* is the residual sum of squares from the regression model, *k* denotes the number of model parameters, and *n* is the sample size. Information criteria tend to penalize complex models, giving preference to simpler models in selection.

Table 3 summarizes the top performance of each selective model and the full model. As shown in the second row of the table, the number of words in a document was the first feature with the highest validity, and thus this feature was involved in the first model. The remaining features are added in turn to the model, according to their significant individual contributions, as described in the second column of Table 3. As shown, when gept1 was added to the model, the results greatly improved; the RMSE of the second model dropped to 0.96 and the correlation rose to 0.82,. This represents a positive contribution for the age of word acquisition. Thus, these results imply that when given a document, the estimated level can be accurately calculated by using the number of words and the proportion of the gept1 word list combined into the regression model.

Based on BIC values, the best model was a combination of the following features: the number of words, gept1, tree height, vq12 and vq13. Our results show that the fifth model in Table 3 had the least difference between the gold truth and the estimated levels: the RMSE is as small as 0.84 and the correlation turned out to be closer to 1 at 0.87. With the other features added, the performance remained steady until the seventh model. After that, the performance began to decrease. This finding suggests that the age of word acquisition and the average complexity of sentence structures are important factors in reading difficulty and should be taken into consideration. Except for the average tree height, the gept1 word list refers to words that users have already learned, while the vq12 and vq13 word lists contain vocabulary that are currently being acquired, corresponding to the specific readers' ability. From these results, we conclude that for non-native readers, previously learned vocabulary, current new vocabulary, and the complexity of sentence structure lead to a successful reading difficulty estimation.

Table 3: Results of the optimal model selection.

| Model | Added Feature | RMSE | *r* | BIC | RSS |
|---|---|---|---|---|---|
| 1 | word_number | 1.25 | 0.67 | 157.50 | 532.87 |
| 2 | gept1 | 0.96 | 0.82 | -20.19 | 311.58 |
| 3 | tree_height | 0.87 | 0.86 | -87.77 | 251.39 |
| 4 | vq12 | 0.85 | 0.86 | -99.52 | 238.79 |
| **5** | **vq13** | **0.84** | **0.87** | **-106.52** | **229.99** |
| 6 | proper_noun | 0.84 | 0.87 | -104.46 | 227.47 |
| 7 | vq15 | 0.84 | 0.87 | -101.14 | 225.80 |
| 8 | vq5 | 0.85 | 0.86 | -97.86 | 224.12 |
| 9 | antecedent | 0.85 | 0.86 | -94.88 | 222.26 |
| 10 | vq11 | 0.85 | 0.86 | -91.40 | 220.74 |
| | all | 1.51 | 0.64 | 121.91 | 208.15 |

To better compare these potential models, the performance of selected models is presented in Figure 1. The upper half of the figure illustrates the RMSE among the models with increased feature numbers, while the lower half of the figure shows the correlation between the models and the ground truth. Initially, the RMSE value decreases and correlation sharply rises as features are added. After identifying the most accurate model, the performance of both measures levels off. This can be seen as a great advantage of model selection, since a small number of identified features achieves a satisfying outcome. Until the Google search result variable was added to the model (the 29th model), the RMSE rapidly increased and the correlation significantly declined. This implies that frequency in a large corpus, such as Google, might not be as useful as the age of word acquisition in reading difficulty estimation. These results reinforce the assumptions of previous studies (Huang, Chang, Sun and Chen, 2011), where the age of word acquisition has more relative importance than frequency within corpora. After the 29th model, the performance fluctuated and worsened, compared to previous models in both measurements. This indicates that performance becomes unstable if the model over-fits. This error may be due to the fact that these models capture idiosyncrasies of the training data rather than generalities.

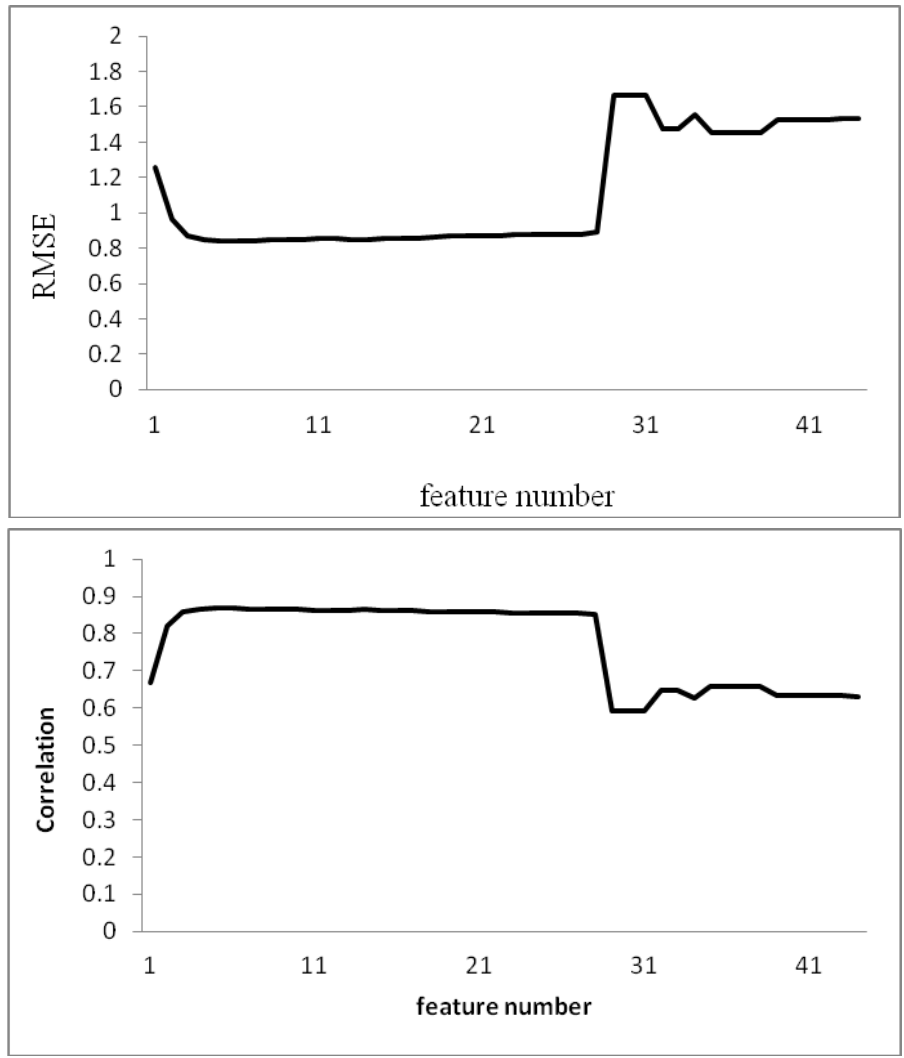

Figure 1: the performance of a selected model.

To understand the impact of feature sets, we also investigated the performance of several regression algorithms with two different feature sets. These algorithms are available in the WEKA package (Bouckaert, Frank, Hall, Holmes, Pfahringer, Reutemann, …and Sonnenburg,, 2010), including Support Vector Regression (SVR; EL-Manzalawy and Honavar, 2005), Sequential Minimal Optimization for Regression (SMOreg; Shevade, Keerthi, Bhattacharyya and Murthy, 2000), Pace Regression (Wang and Witten, 2002), and linear regression. All parameters were used as a default setup. Adopting regression as a reading difficulty model assumes that the output scores are continuous and related with each other.

Table 4 shows the results of these regressions. All methods with only the optimal features outperformed those with all features. This indicates that the optimal feature set could help regression estimates. In addition, the results of the linear regression outperformed those of SVR. This finding is in contrast with Kate et al. (2010), which noted similar performance among regression algorithms. SMOreg, which is a SVR improved by Sequential Minimal Optimization, had the best performance among the models with all features; however, linear regression with optimal features matched the results of SMOreg and Pace Regression. This supports our contribution; when the optimal feature set is identified, the performance among various regression techniques is similar.

Table 4: Comparison of other regression families with the proposed method.

| Method | All Features | | Optimal Features | |
|---|---|---|---|---|
| | RMSE | r | RMSE | r |
| Support Vector Regression (nu-SVR) | 1.66 | 0.22 | 1.24 | 0.60 |
| Support Vector Regression (epsilon-SVR) | 1.65 | 0.25 | 1.42 | 0.59 |
| SMO for Regression (SMOreg) | 1.25 | 0.73 | 0.84 | 0.87 |
| Pace Regression | 1.95 | 0.52 | 0.84 | 0.87 |
| **Linear Regression (Proposed)** | **1.51** | **0.64** | **0.84** | **0.87** |

**Reading Difficulty as Classification**

The proposed model can also be modeled as a multiclass classification, and provide experts with an evaluation that examines in what categorizes certain learning material should be placed. The labels were determined by eight levels ranging from zero to seven: zero represents the documents under the specific readers' ability such as elementary textbooks; seven represents the documents above the specific readers' ability such as college textbooks; the remaining levels are as the same as the semester grades of senior high school. First, the thresholds were found and the estimated levels were assigned to the closest level corresponding to the threshold. The minimum threshold was assigned the minimum value from the training dataset; likewise, the maximum threshold was selected the maximum value from the training dataset.

To understand the performance of the proposed estimation compared with other studies, our experiment also compared the estimated levels within the Flesch Reading Ease (Flesch, 1948), Flesch–Kincaid Grade Level (Kincaid et al., 1975), Coleman-Liau (Coleman and Liau, 1975), Lexile (Stenner, 1996), and the Heilman method (Heilman et al., 2007). The Flesch–Kincaid Grade Level, Flesch Reading Ease, and Coleman-Liau were duplicated, while Lexile and the Heilman method are available online. All of these methods are designed for native readers. In the training phase, the output score generated from each document by those estimations is found, like the procedure of the proposed method, as well as the threshold between each level of the other estimations. During the testing phase, the estimated levels of testing documents were determined using these thresholds.

Not only the RMSE and correlation coefficient, but also accuracy and trend accuracy in direction (TAD) are adopted as measurements in the evaluation. Accuracy is defined as the proportion of the correctness of generated results within the ground truth. The TAD is used in the performance of trend forecasting (Zhang, Jiang, and Li, 2005). The result can be interpreted as the proportion of the same direction between the estimated results and the gold truth. To employ the measurement, the TAD was modified as:

$$TAD = \frac{\sum_{i}^{n}\sum_{j}^{n} D(i,j)}{n \times (n-1)} \times 100\%$$

$$D(i,j) = \begin{cases} 1, \text{ where } (y_i - y_j)(\hat{y}_i - \hat{y}_j) > 0 \\ 0, \text{ otherwise} \end{cases}$$

where $y_i$ is the gold truth, $\hat{y}_i$ is the estimated level, and $n$ is sample size.

Table 5 shows the results between the proposed estimation and other estimations. For accuracy and RMSE, the proposed estimation obviously produced a more accurate reading difficulty prediction than other estimations. When the proposed estimation fails to predict the correct reading difficulty, its error ranges are almost within one grade; by comparison, the error ranges of Flesch–Kincaid Grade Level, the Lexile and the Heilman method were between one to two grades, and Flesch Reading Ease and Coleman–Liau had an even wider error range. Through TAD, the proposed estimation was consistent with the ground truth, although it might tend to predict easy documents with a lower grade and difficult documents with higher grades. In contrast, the results of the other method are fluctuant. In the correlation coefficient, all estimations are positively correlated. The proposed estimation reported a particularly high correlation at *0.87*, whereas the other estimations were at *<0.5*. This suggests that the relationship between the proposed estimation and the ground truth is stronger than others. In summary, existing difficulty estimation methods perform poorly for non-native readers, which may due to the different and insufficient features used.

Table 5: Comparison between the estimations.

| Estimations | RMSE | r | Accuracy | TAD |
| --- | --- | --- | --- | --- |
| Flesch Reading Ease | 2.17 | 0.27 | 0.28 | 0.40 |
| Flesch–Kincaid Grade Level | 1.85 | 0.48 | 0.26 | 0.49 |
| Coleman–Liau | 2.16 | 0.31 | 0.24 | 0.41 |
| Heilmen | 1.84 | 0.41 | 0.26 | 0.43 |
| Lexile | 1.76 | 0.46 | 0.33 | 0.49 |
| **Model 5 = word_number+gept1+tree_height+vq12+vq13** | **1.01** | **0.87** | **0.42** | **0.68** |

**Discussion**

We propose a reading difficulty model based on linear regression. This model not only inherently employed the complexity of lexical and syntactic features, but also newly introduced some meaningful new features such as the age of words and grammatical acquisition, word sense, and coreferences. We further identified the performance of the categorized features and the individual features. Regardless of the categorized results or the individual results, the age of word acquisition consistently exceeded the majority of other features. This suggests that the age of word acquisition as word difficulty is an important characteristic of successful reading difficulty of non-native readers. We also investigated how the combination of features improves reading difficulty estimation based on model selection. We found that the combination of word number, the age of word acquisition and the height of parsing tree had the best performance among all of models. Such an explanation may account for non-native readers, and these features are crucial in understanding the content of a document. Here, the number of words in a document is correlated with its difficulty level. Moreover, the age of word acquisition shows that the higher the proportion of words in a document that a non-native reader already understands, the simpler the document's reading level. Lastly, the height of the parsing tree demonstrates that the fewer sentence structures, the lower the difficulty. We compared the proposed estimations with those of other methods, and our experimental results clearly revealed that our proposed estimations are more suitable for non-native readers, relative to other systems designed for native readers.

As highlighted above, our empirical evaluation demonstrates that the proposed estimation seems to have benefited from the age of word acquisition. In our study, we adopted two organized word lists based on word difficulty to represent this indicator. Although these two expert-designed lists were both available, it is uncertain whether the performance of the model is consistent across a wide combination of word lists. Thus, we conducted an experiment with model selection to analyze the performance of more refined and compatible word lists. The VQ features were specifically more refined, which covered words learned in different grades from a elementary to college. In contrast, the GEPT features only examined language acquisition after junior high school, senior high school and college. These results are presented in Table 6. As noted, even though both VQ and GEPT had slightly lower than proposed model in the RMSE, they were apparently similar to the proposed model in the Correlation. This shows that a model with only one word list is still comparable in performance with models that use two or more lists.

Table 6: Comparison of the model with different word lists.

| Category | Model | RMSE | $r$ | BIC | RSS |
|---|---|---|---|---|---|
| Optimal | 0.0035 * word_number + -5.0912 * gept1 + 17.091 * vq12 + 24.7411 * vq13 + 0.312 * tree_height + 1.0879 | 0.84 | 0.87 | -106.52 | 229.99 |
| VQ | 0.0035 * word_number +-3.7323 * vq5 +11.2876 * vq11 +18.8208 * vq12 +28.3072 * vq13 + 0.3181 * tree_height | 0.87 | 0.84 | -94.94 | 229.93 |
|  | +-2.1742 |  |  |  |  |
| GEPT | 0.0035 * word_number + -3.7106 * gept1 + 6.2536 * gept2 +16.3416 * gept3 + 0.3141 * tree_height +-0.6254 | 0.85 | 0.87 | -94.79 | 233.99 |
| All |  | 1.51 | 0.64 | 121.91 | 208.15 |

In the field of language learning and cognition, constructing an organized word list that tracks word difficulty has been widely investigated. There are two key examples from second language learning environments: in Taiwan, organized word lists during elementary school and high school have been designed by educational organizations; in Japan, words with grading indices are also studied by language learning companies (http://www.alc.co.jp/eng/vocab/svl/index.html). In the field of psychology, the University of Western Australia has developed the MRC Psycholinguistic Database as a word list (Wilson, 2003). In Chall and Dale (1995), the authors also adopted a list of 3,000 basic words and counted the number of words not appearing in the list to estimate reading difficulty. However, such word lists require significant labor to produce, and they sometimes do not exist for languages other than English. Kireyev and Laudauer (2011) have tried to computationally estimate word difficulties. As a new analytical tool, this area requires further study.

As mentioned in Section 1, the model was motivated by providing textbook writers with an estimated document difficulty standard, in order to help them design a textbook. An additional application could apply this work to any reader, especially non-native readers, and help them select suitable articles from among the vast numbers of choices online. Initial work in this area includes Collins-Thompson et al. (2011), where search results were re-ranked by a reading level of each query-related response. Our study found strong predictors of reading difficulty estimation for second language materials. Furthermore, although we use English textbooks from Taiwan in our experiments, our model could also have broad applicability to other languages.

**Conclusion**

Second language learners are an important segment of the global population, however their second language reading skills may perform more poorly compared to native readers, because of their learning experience and processing. In this work, we present an estimation of reading difficulty designed for non-native readers. The proposed estimation is not only based on several meaningful lexical and grammatical features from existing prior research, but also on several novel features, such as the age of words and grammatical acquisition from several sources. In addition, we also proposed features that consider word sense and coreference resolution. These features were extracted and sent to a linear regression model to estimate a reading level of a document.

We applied the model on a data set of senior high school English textbooks designed for Chinese students in Taiwan to learn English as a second language. The empirical results identified the performance of the individual features, and showed that the age of word acquisition plays an important role in the reading difficulty estimation designed for non-native readers. We also investigate how the combination of features improving reading difficulty estimation based on model selection. We found the performance of the combination of word number, the age of word acquisition and the height of parsing tree (*RMSE=0.84, r=0.87*) outperformance the rest of models. This may be viewed as a validation of age of word acquisition in reading difficulty estimations for non-native readers. After finding the optimal model, we compared the proposed estimations with those of other methods. Our proposed estimations achieved the highest performance among all measurements, and were higher than other state-of-the-art estimations.

In the future, we will integrate the proposed estimation into the AutoQuiz (http://autoquiz.iis.sinica.edu.tw/) project and provide second language learners with personalized learning materials.